\documentclass[runningheads]{llncs}
\usepackage{graphicx}

\usepackage{tikz}
\usepackage{comment}
\usepackage{amsmath,amssymb} 
\usepackage{color}

\usepackage[accsupp]{axessibility}  

\usepackage{booktabs}    

\begin{document}
\pagestyle{headings}
\mainmatter
\def\ECCVSubNumber{100}  

\title{ELSR: Extreme Low-Power Super Resolution Network For Mobile Devices} 

\author{Tianyu Xu, Zhuang Jia, Yijian Zhang, Long Bao, Heng Sun
}
\institute{Video Algorithm Group, Camera Department, Xiaomi Inc.\\
\email{\{xutianyu,jiazhuang,zhangyijian,baolong,sunheng3\}@xiaomi.com}
}

\maketitle

\begin{abstract}

With the popularity of mobile devices, e.g., smartphone and wearable devices, lighter and faster model is crucial for the application of video super resolution.
However, most previous lightweight models tend to concentrate on reducing lantency of model inference on desktop GPU, which may be not energy efficient in current mobile devices.
In this paper, we proposed \textbf{Extreme Low-Power Super Resolution (ELSR)} network which only consumes a small amount of energy in mobile devices. Pretraining and finetuning methods are applied to boost the performance of the extremely tiny model.
Extensive experiments show that our method achieves a excellent balance between restoration quality and power consumption.
Finally, we achieve a competitive score of 90.9 with PSNR 27.34 dB and power 0.09 W/30FPS on the target MediaTek Dimensity 9000 plantform, ranking 1st place in the Mobile AI \& AIM 2022 Real-Time Video Super-Resolution Challenge.

\end{abstract}

\section{Introduction}

Video Super-resolution (VSR) task aims to restore high-resolution (HR) videos from  low-resolution (LR) videos. In contrast to single image super-resolution (SISR) methods, which can only rely on natural image priors for estimation of high resolution details, VSR can also exploit temporal information for improved recovery of details in each frame.
In the past few year, a series of CNN-based methods have achieved state-of-the-art results in various computer vision tasks, including super resolution. At the same time, the popularity of intelligent mobile devices such as smartphone and wearable devices drives high demand for VSR.
Although mobile systems in a chip (SoC) have achieved excellent hardware acceleration performance after continuous development, deploying a CNN-based model which is fast and energy-friendly in mobile devices is still a challenging task. 

To promote the development of mobile VSR technology, Mobile AI \& AIM 2022 Workshop holds the Real-Time Video Super-Resolution challenge which aims to super-resolve an input LR video to a corresponding HR video in the spatial domain with a upsampleing factor 4, using TFLite model running on mobile devices.
Participants are asked to balance the restoration quality and the power consumption of 
the submitted model on the target mobile platform. The power consumption is measured based MediaTek NeuroPilot.
In this paper, we provide \textbf{Extrem Low-Power Super Resolution (ELSR)} for tackling VSR problem with low-power and real-time restrictions.
Extensive experiments have proven the efficiency of proposed solution, and we won the 1st place out of 11 teams in final test phase of the challenge with score of 90.9.

\section{Related Works}

\subsection{Video Super-Resolution}

As one of the most popular topics in computer vision, Super-Resolution has drawn a lot of attention from researchers. With the development of deep learning \cite{ref1}, methods based on convolution network show great advantages over traditional algorithms such as bilinear and bicubic unsampling. The pioneer work was done by Dong et al. \cite{ref2}, who proposed SRCNN for image super-resolution and achieve outstanding performance compared to previous works. VDSR \cite{ref3} introduces residual connection and increase network depth which with 20 layers, achieving significant improvement in restoration quality. ESPCN \cite{ref4} proposed pixel shuffle operation, and added it at the end of network to directly upsample to the size of HR image from LR feature maps. 

VSR can be viewed as a generalization of image super-resolution. Videos additionally provide temporal information among frames, which can be explored to improve restoration quality. One of the representative work is FRVSR \cite{ref5} which first produces a super-resolved HR output by concatenating the SR output of network from previous LR input and the current LR frame, which obtains better result compared with single frame methods.
Different from FRVSR that only utilize one previous frame to help reconstruct current
LR frame, EDVR \cite{ref6} adopts sliding windows to obtain more temporal information for improving restoration quality. EDVR obtains aligned feature by introducing coarse-to-fine deformable alignment and a new spatio-temporal attention fusion module. As an integrator of video super-resolution models, BasicVSR \cite{ref7} and BasicVSR++ \cite{ref8} proposed a baseline VSR model composed of propagation, alignment, aggregation and upsampling. By conducting bidirectional propagation and deformable convolution for feature-level alignment, BasicVSR series has achieved excellent performance in VSR task.

\subsection{Efficient Networks}

Besides the manual design of deep neural networks, there are a plethora of works that try to improve the effectiveness and efficiency of deep neural networks via network pruning, low-rank filter decomposition, net quantization, neural architecture search and knowledge distillation etc. Among those network compression works, a couple of them have been successfully applied to image SR. However, it is still a challenging topic for mobile device SR because of the limit of computing resources and restrictions on power consumption. IMDN \cite{ref9} proposed a multi-branch distillation network structure to reduce model parameters and speed up running time. RFDN \cite{ref10} further modified model structure, replace channel spilt with $1\times1$ convolution and proposed ESA module to improve restoration quality by introducing attention mechanism. However, the above model still measures lantency and power on the desktop devices which is not suitable for mobile devices.

At the same time, some of the researchers try to deploy deep learning super-resolution model to real-time mobile devices. Edge-SR \cite{ref11} is proposed to super-resolve 4K resolution video streams to 8K resolution video streams in television using only one convolution layer. Under the same parameters, Edge-SR acquires better visual quality, and have less parameters under the same subjective quality. Moreover, in MobileOne \cite{ref12} is proposed as lightweight backbone designed for mobile devices, which utilize re-parameterize tricks to speed up inference time.

Inspired by the above works, we proposed a mobile-friendly VSR model which has very few parameters and FLOPs and can run in real time on the device with low power consumption.

\section{Method}

In this section, we will introduce basic architecture of our model, and provide a a detailed illustration about the design principle of our model by comparing different operators and modules.

\subsection{Baseline}

The core idea of our method is to design a mobile friendly network which consumes as little energy as possible, so we discord some complex operations such as optical flow, multi-frame feature alignment, and start our experiment from single frame baselines.

\begin{table}[htbp]
\setlength{\tabcolsep}{4mm}{ } 
\centering
\begin{tabular}{lclcccc}
\toprule
model & nb & nf & bc & power & PSNR/dB & score\\
\midrule
RFDN & 4 & 24 & 4 & 0.69 & 28.11 & 62.16 \\
PlainSR & 4 & 24 & 3 & 0.49 & 28.08 & 72.11 \\
w/o ESA & 4 & 20 & 3 & 0.44 & 28.04 & 74.54 \\
w ESA   & 4 & 20 & 3 & 0.59 & 28.13 & 67.20 \\
\bottomrule
\end{tabular}
\vspace{1mm}
\caption{Comparison of different model module. nb, nf refer to number of blocks, feature map channels, number of conv layer in each block, and bc refers to number of conv layers for plain architecture and number of branches in RFDN.}
\label{tab_module}
\end{table}

Currently, the lightweight super-resolution models tend to have a multi-branch structure (e.g. IMDN, RFDN). However, as shown above Table \ref{tab_module}, the multi-branch distillation structure can show significant increase in energy consumption while also get a slight increase in PSNR compared with the plain convolutional network of similar parameters. For this reason, we abandoned further exploration on multi-branch network architectures, and focused on plain convolutional SR networks.

Attention mechanism \cite{ref13} is also an effective way to improve the performance of SR models. We also tested some commonly used attention modules such as ESA, CCA and PA \cite{ref14}. Experimental results show that though attention modules bring performance improvement, the extra energy consumption introduced is still unacceptable. As shown in Table 1, ESA module can boost performance by about 0.09dB, but it also increase power consumption by about 0.15W resulting in a lower score.
As a result, we use plain serial network structure without bells and whistles, and a skip connection transfers the input features directly to the latter conv layers to benefit the training and compensate the loss of information. Residuals are used in many lightweight models for super-resolution, but as the add operation of former and latter feature maps will bring more FLOPs, so we use only one residual pathway in the final model.

\begin{figure}[h]
\centering
\includegraphics[width=12cm]{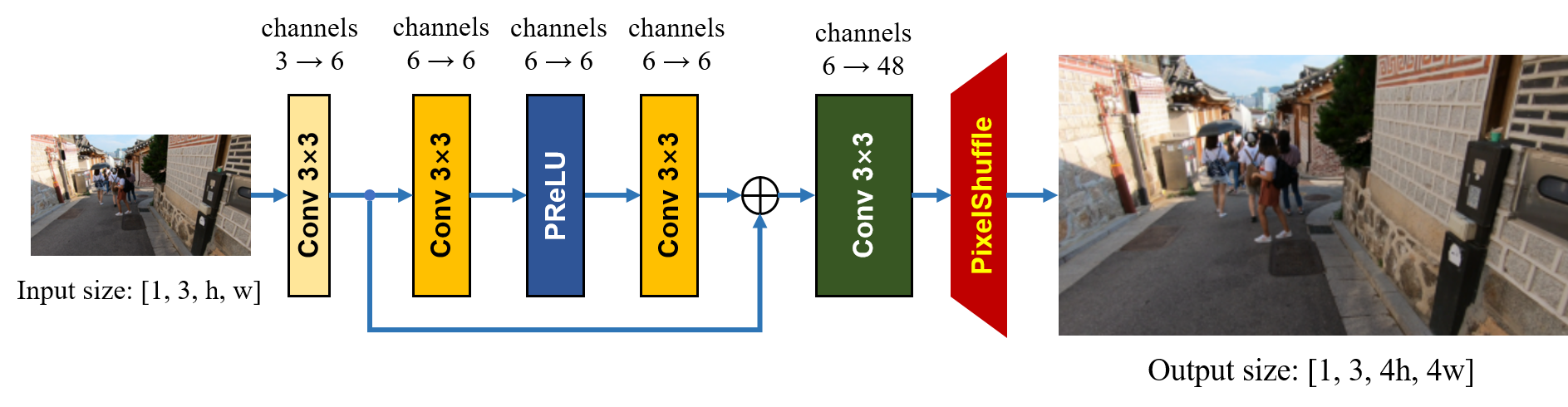}
\caption{Architecture of used network for the challenge}
\label{fig1}
\end{figure}

Finally, we obtain a simple yet effective network structure with single frame input (as shown in Fig.\ref{fig1}) which only have 6 layers, of which only 5 have learnable parameters, including 4 Conv layers and a PReLU activation layer. Pixel-Shuffle operation (also known as $depth2space$) is used at last to upscale the size of output without introducing more calculation. The intermediate feature channels are all set to 6.

\subsection{Activation \& Loss Function}

\begin{figure}[h]
\centering
\includegraphics[width=8cm]{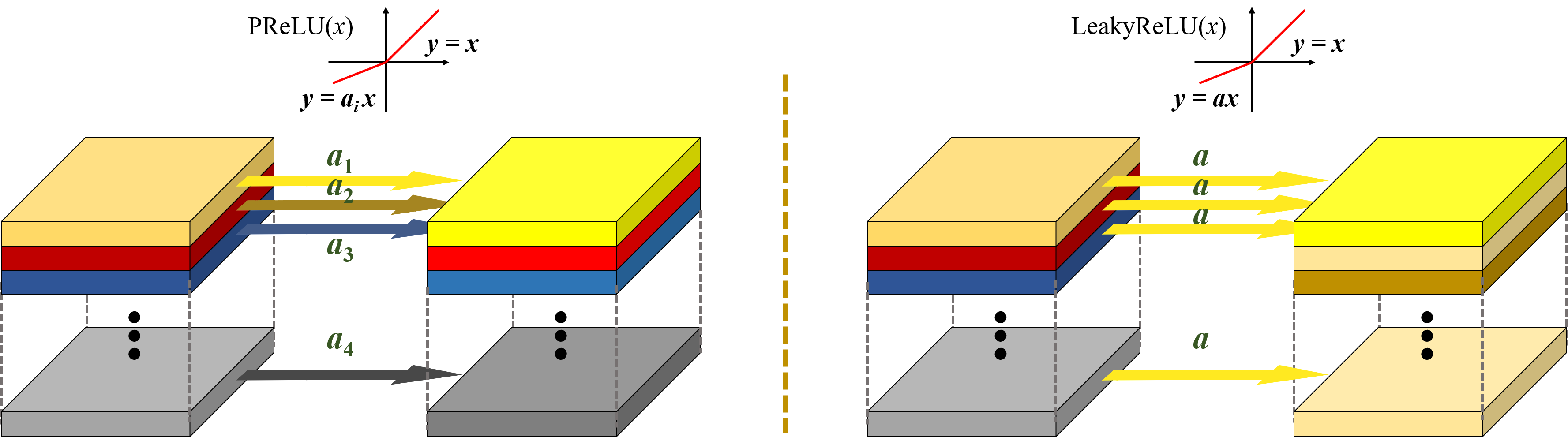}
\caption{PReLU has more capability of representation compared with LeakyReLU}
\label{fig_prelu}
\end{figure}

For the choice of activation function, we compared the commonly used ReLU, Leaky ReLU and Parametric ReLU (PReLU) activations (Tab.\ref{tab_prelu}), and the results imply that the PReLU can boost the performance by about 0.05 dB due to its higher flexibility (Fig.\ref{fig_prelu}). At the same time, PReLU op has nearly the same power consumption as ReLU / LeakyReLU, thus is suitable for our low-power model design.

\begin{table}[htbp]
\setlength{\tabcolsep}{6mm}{ } 
\centering
\begin{tabular}{lclccc}
\toprule
id & network settings & power & PSNR/dB & score\\
\midrule
1 & 4, 12, 3, relu  & 0.25 & 27.84 & 83.71 \\
2 & 4, 12, 3, prelu & 0.24 & 27.91 & 84.33 \\
\bottomrule
\end{tabular}
\vspace{1mm}
\caption{Comparison of two different activation functions. Network settings same as above.}
\label{tab_prelu}
\end{table}

We utilize L1 and L2 loss (i.e. MSE loss) function separately since we employ a multi-stage training approach. Firstly, we use L1 loss function for pre-training the model to achieve better convergence, and then the previous model is finetuned using L2 loss to reach higher PSNR score.

\section{Experiments}
\subsection{Dataset}
 
Our model is trained on REDS \cite{ref15} dataset, which have a large diversity of contents and dynamic scenes. It is widely used in video super-resolution and video frame-interpolation task. REDS dataset consists of 300 video sequences with each 100 frames of 720 $\times$ 1280 resolution, and the organizer provides the corresponding low-resolution video sequences by dowmsampling scale $\times4$. At the same time, We generated low-resolution video sequences which downsampling scale $\times2$ by ourselves using matlab bicubic function for model pretraining. In this challenge, the dataset is divided into 240 sequences for training, 30 sequences for validation, and the left 30 sequences for testing.
 
\subsection{Implementation Details}

We adopt REDS dataset including $\times2$ and $\times4$ low-resolution video sequences for training where $\times2$ data is used for model pre-training. In order to accelerate the model training process, all training video frames are randomly cropped to patches.

\begin{figure}[h]
\centering
\includegraphics[width=10cm]{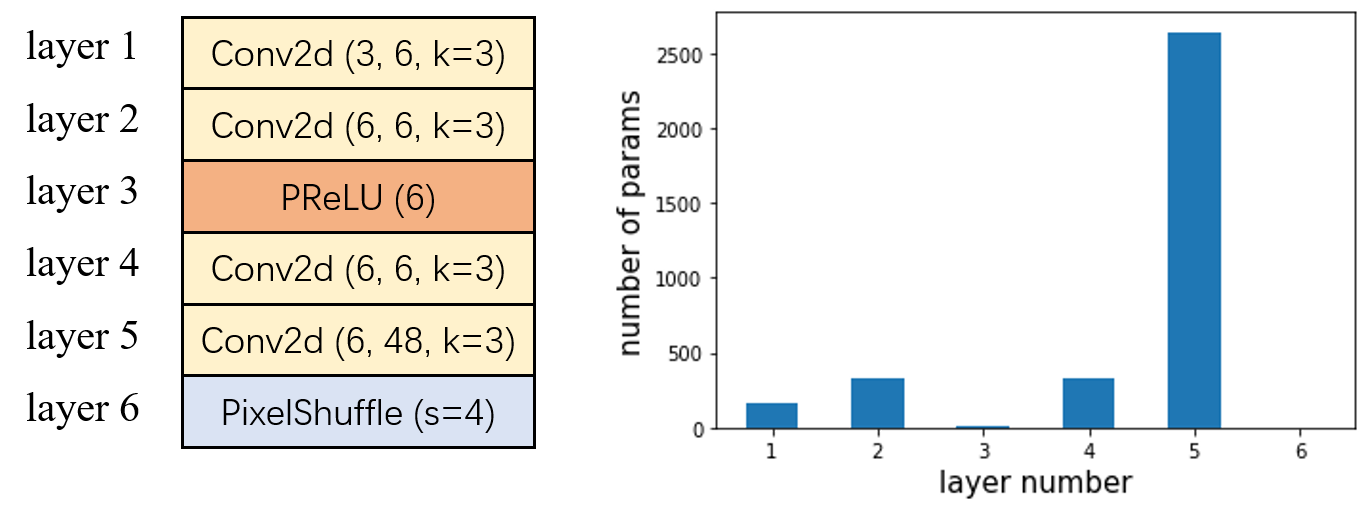}
\caption{Layer parameter number distribution in ELSR model}
\label{fig_layer_weight}
\end{figure}

Pre-training on $\times2$ task with the same training dataset for $\times4$ task is a commonly used strategy in super-resolution tasks. We adopt this strategy with some modification due to the extreme tiny structure of our network. For $\times2$ pre-training, the last conv layer with weight of size $[48,8,3,3]$ will be replaced with $[12,8,3,3]$ for $\times4$ finetuning. As a common case, after the $\times2$ model training is finished, the weights are loaded to $\times4$ model except for the last layer which is incompatible in size. However, in our extremely small model, the weight of last layer occupies more percentage in all parameters compared with normal or large SR models \ref{fig_layer_weight}.

\begin{figure}[h]
\centering
\includegraphics[width=10cm]{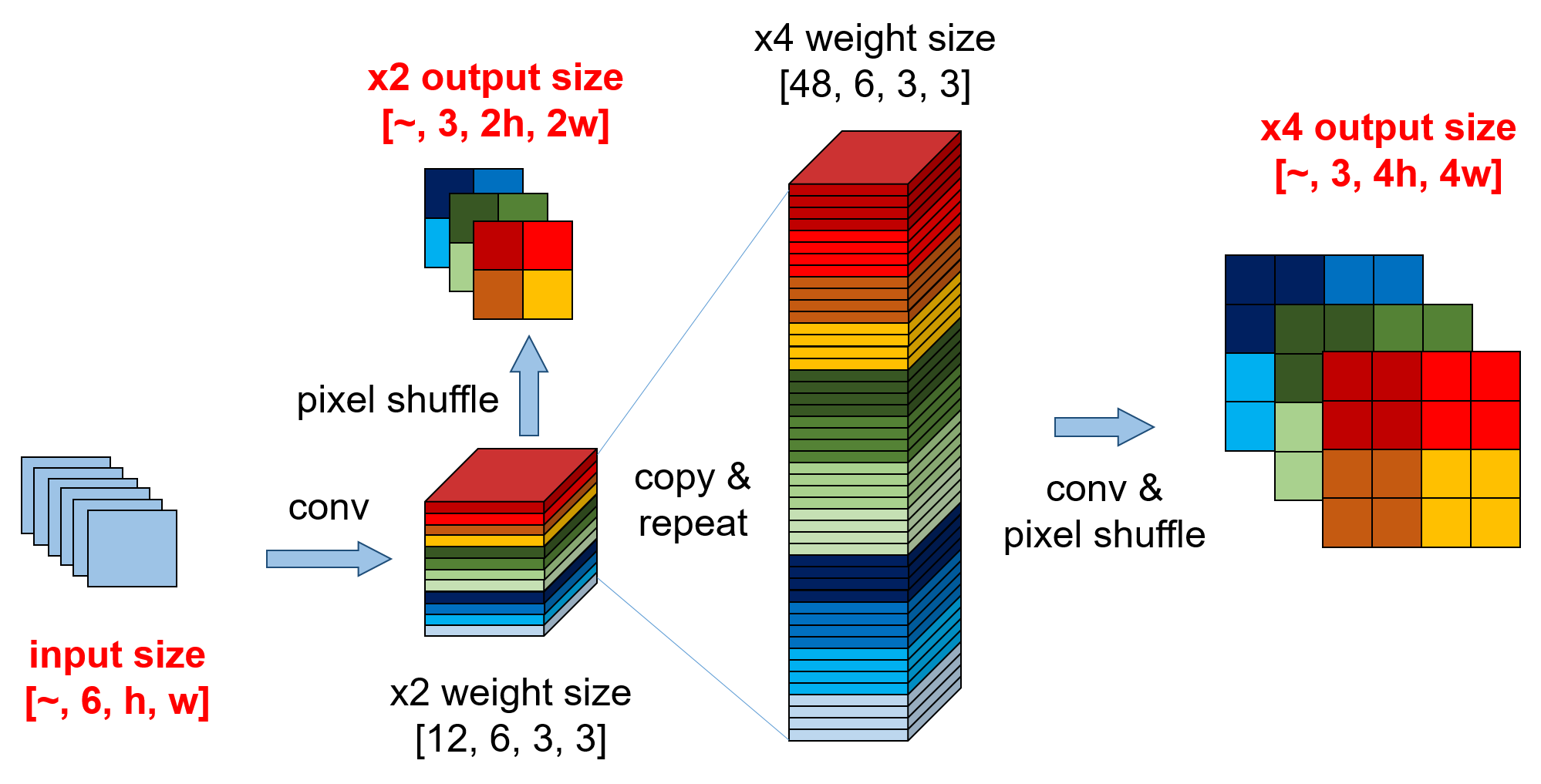}
\caption{Repeat last conv layer for adapt $\times2$ pretrained weight to $\times4$ training}
\label{fig2}
\end{figure}

Considering the operation of pixel shuffle, we keep the weight from last layer in $\times2$ model and repeat the weight 4 times along channel dimension, complying with the spatial position of the output after pixel shuffle (Fig.\ref{fig2}). After this process, the output of $\times4$ model loaded with $\times2$ weights becomes the nearest interpolation of the $\times2$ output, which accelerates the convergence in training $\times4$ model.

The model is trained for six stages, and the detailed training configurations are as follows: 

(I) Pre-train a $\times2$ model from scratch as stated above, with batch size 64 and patch size (HR) 256 for 500k iterations. The model is optimized using L1 loss with Adam optimizer, with initial learning rate of $5e-4$, and the learning rate decreases by 0.5 in 200k, 400k iterations respectively.

(II) Train $\times4$ model using the pre-trained weights from $\times2$ model. Batch size is 64 and patch size is 256, total iteration number is 500k. The model is optimized using L1 loss with Adam optimizer, with initial learning rate of $5e-5$, and the learning rate decreases by 0.5 in 100k, 300k and 450k iterations respectively.

(III) Continue to finetune the model from Stage (II) using L1 loss, batch size 64 and patch size 256, and optimizer type stay the same. The model is finetuned for 300k iterations, with initial learning rate of $2e-4$ which decreases by 0.5 in 200k iterations.

(IV) Finetune the model from Stage (III) using MSE loss, with patch size 256 and batch size 64. This stage takes 1000k iterations with learning rate is $2e-4$, and the learning rate decreases by 0.5 in 300k, 600k and 900k iterations respectively.

(V) Continue to finetune the model from Stage (IV) using MSE loss, with patch size 512 and batch size 64. This stage takes 500k iterations with learning rate is $2e-4$, and the learning rate decreases by 0.5 in 100k, 200k, 300k and 400k iterations respectively.

(VI) Finally, finetune the model from Stage (V) using MSE loss, with patch size 640 and batch size 64. This stage takes 50k iterations with learning rate is $2e-5$.

\subsection{Results}

\begin{figure}[h]
\centering
\includegraphics[width=12cm]{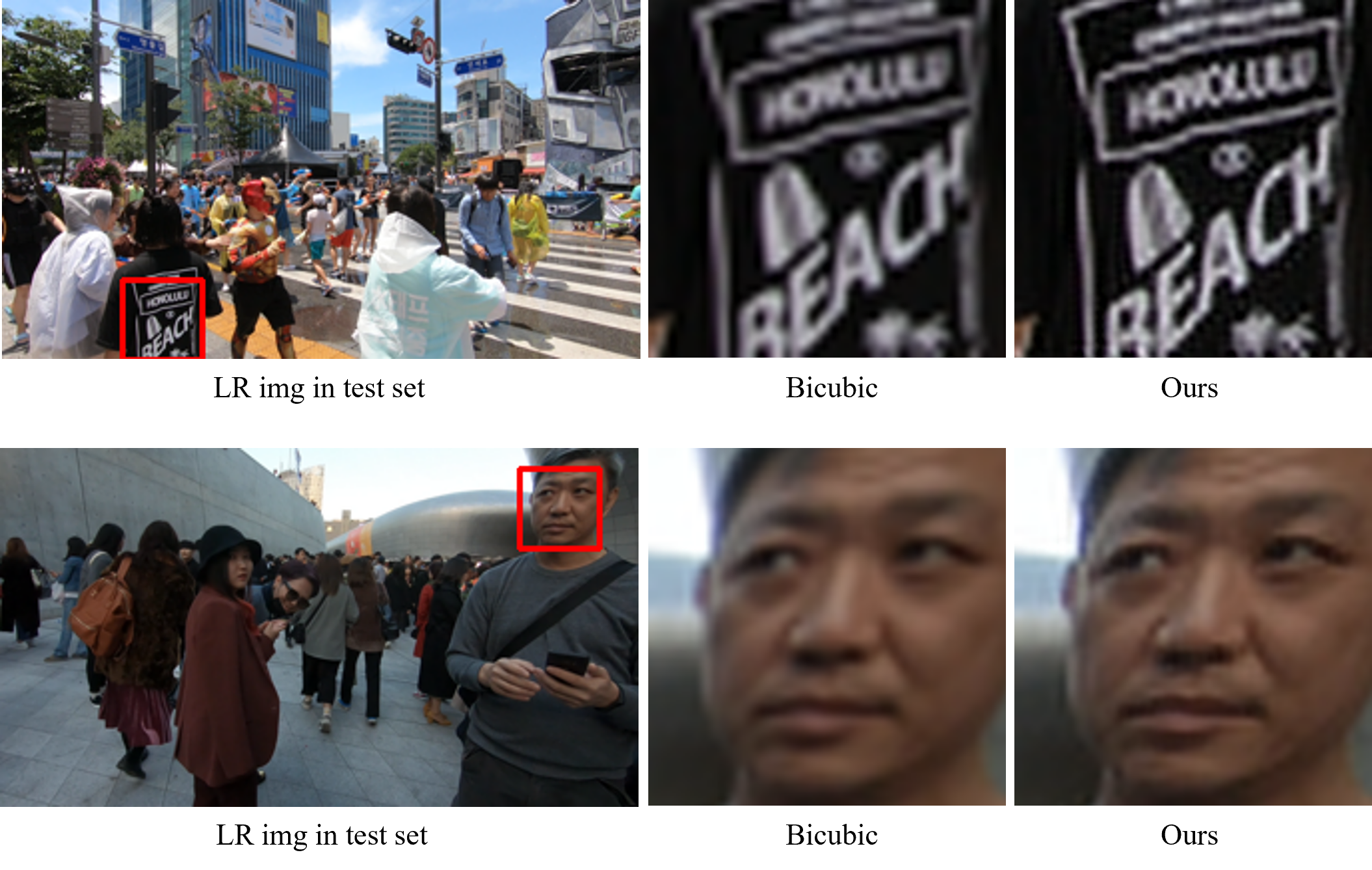}
\caption{Visual comparison of frames in REDS testset, for scale factor 4}
\label{fig3}
\end{figure}

In this challenge, REDS test-set is used to evaluate the quality of the reconstructed results, and the power consumption of the model on the actual MediaTek 9000 platform is tested by organizers. The top three results are shown in Table \ref{tab_result}. As shown in the leaderboard, our solution achieves a competitive result of 90.9 on challenge test set where the PSNR is 27.34 dB and the power consumption is 0.09W/30FPS, ranking 1st place in the Mobile AI \& AIM 2022 real-time VSR challenge. Notably, our method has the lowest power consumption among all participants. Fig \ref{fig3} shows the comparison of our results with bicubic. The result shows our method has more clear edges and details compared with traditional bicubic interpolation algorithm while is power efficient.

\begin{table}[htbp]
\setlength{\tabcolsep}{6mm}{ } 
\centering
\begin{tabular}{lclccc}
\toprule
Team & PSNR/dB & Power & Score\\
\midrule
\textbf{1 (Ours)} & 27.24 & \textbf{0.09} & \textbf{90.9} \\
2 & 27.52 & 0.10 & 90.7 \\
3 & 27.34 & 0.20 & 85.4 \\
\bottomrule
\end{tabular}
\vspace{1mm}
\caption{Mobile AI \& AIM 2022 challenge results}
\label{tab_result}
\end{table}

\section{Conclusions}

In this paper, we proposed Extreme Low-Power Super-Resolution Network (ELSR) which can run in real-time on the mobile devices with very low power consumption. Our network contains only $3\times3$ convolution, PReLU activation and pixel shuffle operation, which is mobile friendly. Experiments with module ablations and challenge results have proven the effectiveness and efficiency of our method.




%
%


\end{document}